\documentclass{article}

\usepackage{arxiv}

\usepackage{multirow}
\usepackage{array}
\usepackage[ruled,vlined]{algorithm2e}

\usepackage[utf8]{inputenc} 
\usepackage[T1]{fontenc}    
\usepackage{hyperref} 
\usepackage{scalerel,stackengine}
\stackMath
\newcommand\reallywidehat[1]{%
\savestack{\tmpbox}{\stretchto{%
  \scaleto{%
    \scalerel*[\widthof{\ensuremath{#1}}]{\kern-.6pt\bigwedge\kern-.6pt}%
    {\rule[-\textheight/2]{1ex}{\textheight}}
  }{\textheight}%
}{0.5ex}}%
\stackon[1pt]{#1}{\tmpbox}%
}

\usepackage{graphicx}
\usepackage[export]{adjustbox}
\usepackage{xcolor}
\newcommand\Tstrut{\rule{0pt}{2.1ex}}       
\newcommand\Bstrut{\rule[-0.9ex]{0pt}{0pt}} 
\newcommand{\TBstrut}{\Tstrut\Bstrut} %

\usepackage{subfigure}
\usepackage{colortbl}
\usepackage{paralist}
\usepackage{bbm}
\usepackage{url}            
\usepackage{booktabs}       
\usepackage{amsfonts}       
\usepackage{nicefrac}       
\usepackage{microtype}      
\usepackage{lipsum}	
\usepackage{xcolor}
\newcommand{\fooAlter}{\hspace{0pt}\textcolor{blue}{$\bullet$} \hspace{5pt}}

\title{Discovering Bilingual Lexicons in Polyglot Word Embeddings}

\author{
Ashiqur R. KhudaBukhsh\thanks{Ashiqur R. KhudaBukhsh and Shriphani Palakodety  are equal contribution first authors.} \\
  Carnegie Mellon University \\
  \texttt{akhudabu@cs.cmu.edu} \\
  \And
  Shriphani Palakodety$^*$\\
  Onai\\
  \texttt{spalakod@onai.com} \\
\And
Tom M. Mitchell \\
  Carnegie Mellon University\\
  \texttt{tom.mitchell@cs.cmu.edu} \\
}

\begin{document}
\maketitle

\begin{abstract}

Bilingual lexicons and phrase tables are critical resources for modern Machine Translation systems. Although recent results show that without any seed lexicon or parallel data, highly accurate bilingual lexicons can be learned using unsupervised  methods, such methods rely on existence of large, clean monolingual corpora. In this work, we utilize a single Skip-gram model trained on a multilingual corpus yielding polyglot word embeddings, and present a novel finding that a surprisingly simple constrained nearest neighbor sampling technique in this embedding space can retrieve bilingual lexicons, even in harsh social media data sets predominantly written in English and Romanized Hindi and often exhibiting code switching. Our method does not require monolingual corpora, seed lexicons, or any other such resources. Additionally, across three European language pairs, we observe that polyglot word embeddings indeed learn a rich semantic representation of words and substantial bilingual lexicons can be retrieved using our constrained nearest neighbor sampling. We investigate potential reasons and downstream applications in settings spanning both clean texts and noisy social media data sets, and in both resource-rich and under-resourced language pairs.

\end{abstract}


\section{Introduction}
Bilingual lexicons have been a key building block for effective Machine Translation systems for the last few decades~\cite{rapp-1995-identifying, smith2017offline, conneau2017word}. Several prominent recent lines of work on bilingual lexicon induction~\cite{mikolov2013exploiting, dinu-baroni-2014-make, gouws2015bilbowa, luong-etal-2015-bilingual, coulmance-etal-2015-trans,smith2017offline, conneau2017word} involve training embeddings on two monolingual corpora and learning a mapping between the two embeddings. 

In this widely studied NLP problem, primary focus areas are (1) methods (e.g., joint training~\cite{gouws2015bilbowa} versus separate training~\cite{mikolov2013efficient}; incorporating adversarial approaches~\cite{miceli-barone-2016-towards, zhang-etal-2017-adversarial, conneau2017word}) (2) performance and (3) alleviating resource requirements (e.g., seed lexicon, parallel data). While recent successes with unsupervised methods demonstrate that effective lexicons can be learned without any seed lexicon or parallel data~\cite{conneau2017word}, to the best of our knowledge, no prior work has focused on an extreme condition of learning bilingual lexicons from noisy, multilingual social media data using unsupervised methods. In the context of Romanized Hindi, an expression of Hindi primarily observed in social media, this in fact is a real world challenge. With a combined language base of more than 500 million Hindi speakers in India and Pakistan, and prior studies reporting that more than 90\% of Indian language texts found on the web are Romanized~\cite{gella2014ye}, rich bilingual lexicons would help in the under-explored task of analyzing Romanized Hindi web content.

In this paper, we describe a surprisingly simple technique to retrieve bilingual lexicons. We find that by (1) training a single Skip-gram model on the whole multilingual corpus, and (2) conducting a constrained nearest neighbor sampling on the resulting polyglot word embedding space for a given source word, we can retrieve substantial bilingual lexicons even in harsh social media multilingual data sets. On two data sets found in the literature~\cite{IndPak,election} and one data set first introduced in this paper, we demonstrate that without any parallel data, monolingual corpora or seed lexicon, our method retrieves substantial lexicons. We then demonstrate practical benefits of our bilingual lexicon through a cross-lingual sampling task defined in~\cite{khudabukhsh2020harnessing}. On an important task of detecting peace-seeking, hostility-diffusing user generated web content from heated political discussions between nuclear adversaries, dubbed \emph{hope speech} detection~\cite{IndPak}, we present a purely unsupervised cross-lingual sampling method that detects \emph{hope speech} in Romanized Hindi. Our method obtains a 45\% improvement over previously reported results. 

While our performance numbers are encouraging, we see our paper primarily as a discovery paper where we are intrigued by the observation that polyglot Skip-gram embeddings trained on a multilingual corpus learn a rich semantic representation that captures word meanings across languages. We devote a considerable part of this paper in understanding the possible reasons  and investigating the generality of our approach. Our experiments reveal that our method can retrieve bilingual lexicons even across European language pairs. 


\begin{figure}[t]
\centering
\subfigure[\emph{en}-\emph{hi}$_e$ polyglot Skip-gram embeddings.]{%
    \includegraphics[width = 0.45 \textwidth,height = 0.27 \textwidth]{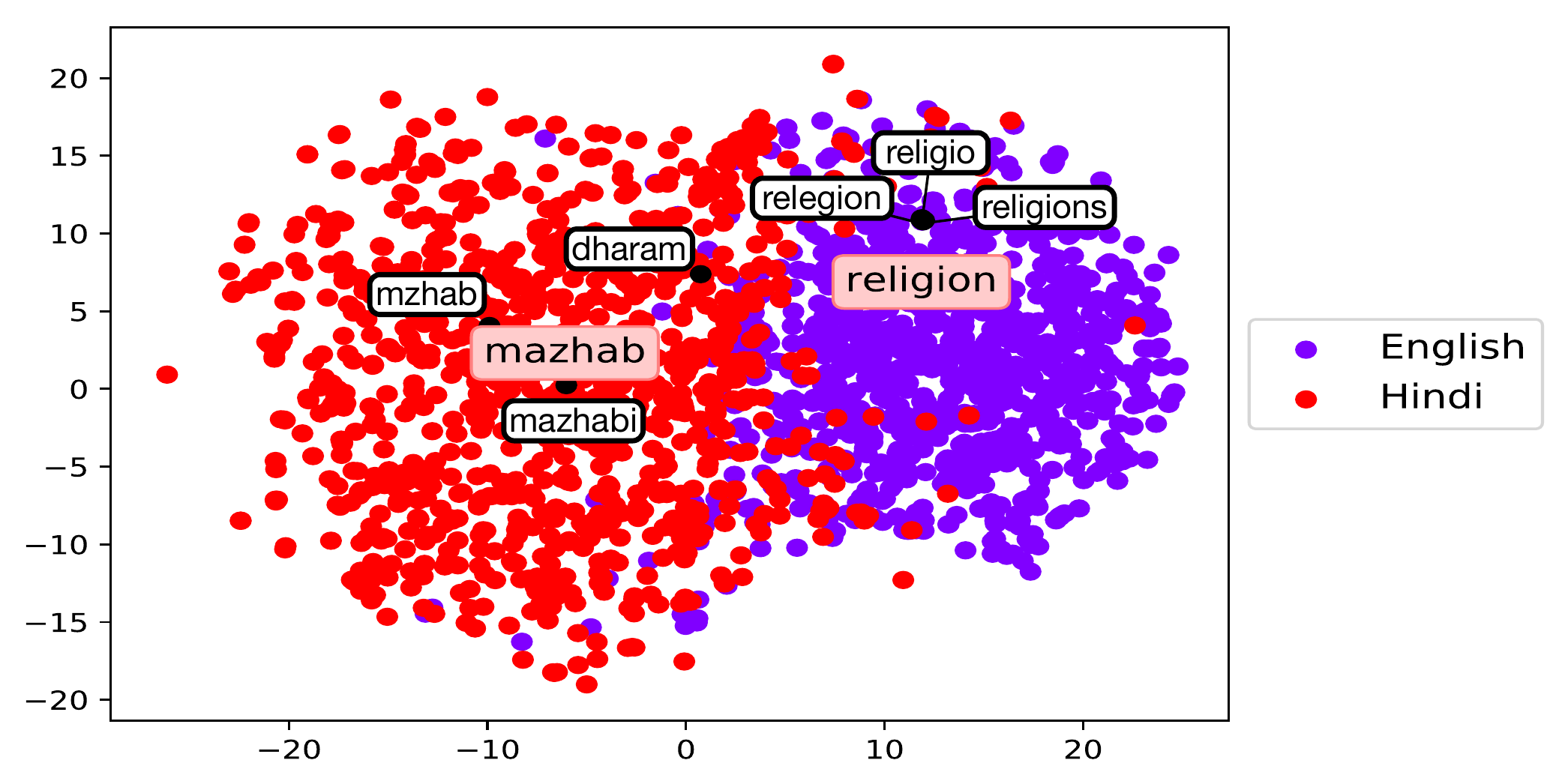}
    \label{fig:hindi}
}
\subfigure[\emph{en}-\emph{es} Skip-gram polyglot embeddings.]{%
    \includegraphics[height = 0.27 \textwidth]{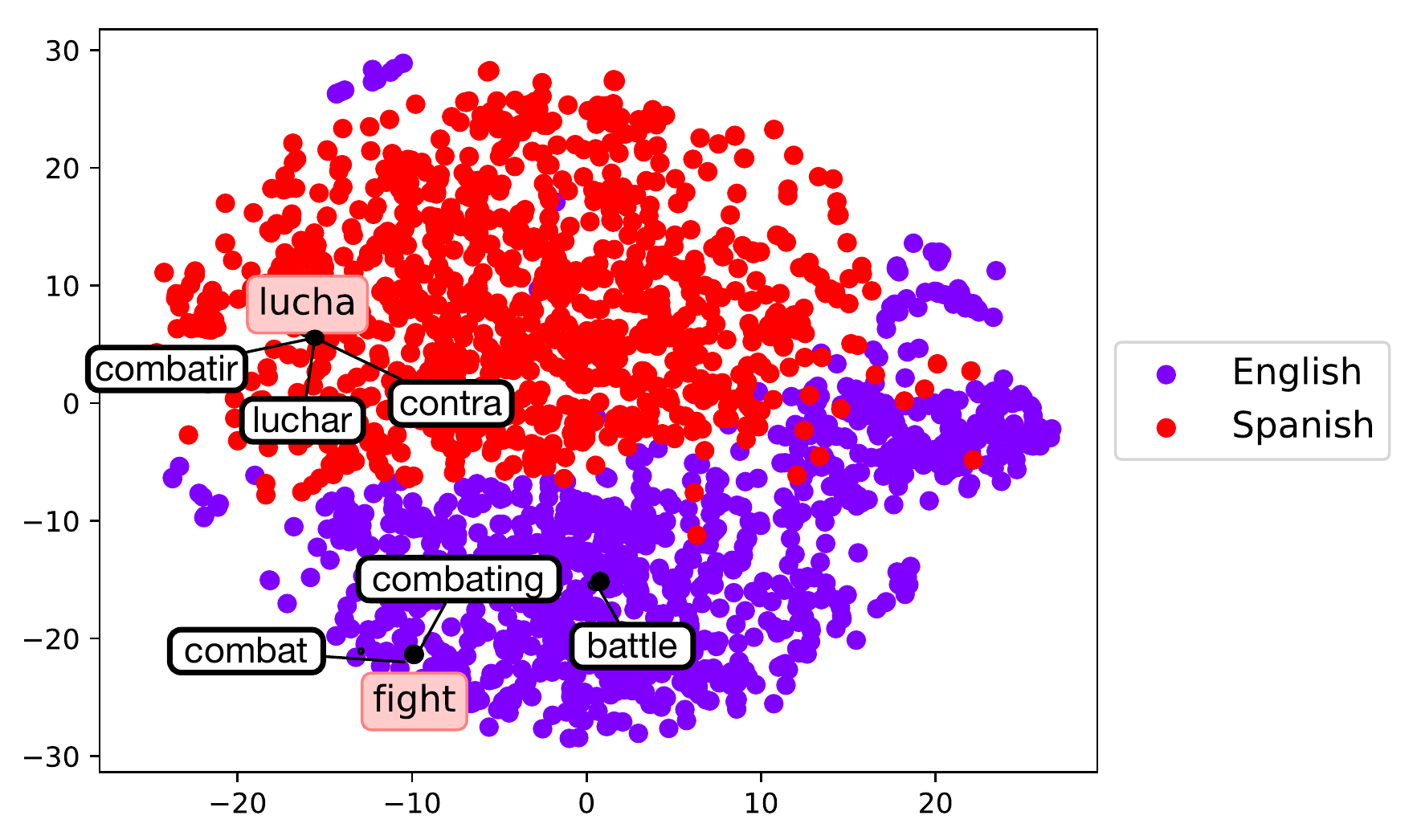}
    \label{fig:spanish}
}

\caption{
A t-SNE~\cite{maaten2008visualizing} 2D visualization of Skip-gram polyglot embeddings trained on (a) a data set of 2.04 million YouTube comments relevant to the 2019 India-Pakistan conflict~\cite{IndPak} and (b) \emph{en}-\emph{es} Europarl corpus~\cite{koehn2005europarl}. In Figure~\ref{fig:hindi}, three nearest neighbors (using cosine distance)  of the Hindi word \emph{mazhab} ``religion'' and the English word \emph{religion} are highlighted. In Figure~\ref{fig:spanish}, three nearest neighbors (using cosine distance)  of the Spanish word \emph{lucha} ``fight'' and the English word \emph{fight} are highlighted. These results indicate that naive nearest neighbor sampling of a word yields other words with similar meaning in the same language. 
}
\label{fig:example}
\end{figure}

\noindent\textbf{\emph{Contributions}}: In this paper, we \textcolor{blue}{(1)} Construct a simple yet capable method to mine substantial bilingual lexicons from noisy social media corpora. \textcolor{blue}{(2)} Focus on a poorly resourced but extremely prevalent language pair: English-Romanized Hindi. \textcolor{blue}{(3)} Release the resulting lexicons of 1,100 word pairs. \textcolor{blue}{(4)} Provide a purely unsupervised cross-lingual sampling technique for an important humanitarian domain.  \textcolor{blue}{(5)} Provide insights into our finding on polyglot word embeddings with a broad study across several language pairs.

\section{Our Approach}

The Skip-gram objective predicts an input word's context~\cite{mikolov2013efficient}. Nearest neighbor sampling in the resulting embedding space yields (syntactically or semantically) similar words; subtle semantic relationships like word-associations (e.g.,  \emph{big}:\emph{bigger}::\emph{quick}:\emph{quicker} or \emph{France}:\emph{Paris}::\emph{Italy}:\emph{Rome}) are also revealed. When two separate Skip-gram models are trained on two monolingual corpora, it is observed that the trained embeddings exhibit isomorphism~\cite{mikolov2013exploiting} and using a seed lexicon, a linear projection can be learned to construct a bilingual lexicon. 
 
\textbf{\emph{What happens when we train a single Skip-gram model on a bilingual corpus ?}}
Intuitively, sampling a word's nearest neighbors in a polyglot embedding space trained on a bilingual corpus would still yield semantically similar words in the same language of the probe word. For instance, as presented in Figure~\ref{fig:spanish}, in  Skip-gram polyglot word-embeddings trained on an English-Spanish Europarl data set~\cite{koehn2005europarl}, nearest neighbors of the Spanish (\emph{es}) word \emph{lucha}, are \emph{contra}, \emph{luchar} and \emph{combatir}; the English (\emph{en}) word \emph{fight} are \emph{fighting}, \emph{combating} and \emph{combat}. Or, in Skip-gram polyglot word-embeddings trained on a noisy social media English-Romanized Hindi corpus of YouTube video comments relevant to the 2019 India-Pakistan conflict~\cite{IndPak}, we notice that nearest neighbors of a word include spelling variations and incorrect spellings of a word along with words with similar meanings. 

Do these polyglot word embeddings capture richer semantic structures that tell us \emph{lucha} and \emph{fight} or \emph{religion} and \emph{mazhab} are related, in fact, synonymous to each other? In this paper, we describe a surprisingly simple constrained nearest neighbor sampling that allows us to retrieve substantial bilingual lexicon that requires (1) no seed lexicon (2) no parallel data and even (3) no monolingual corpora. Recent analyses of polyglot embedding spaces have shown that tokens and documents are grouped together based on their language - i.e., the monolingual components are grouped together. A resulting algorithm $\reallywidehat{\mathcal{L}}_{\emph{polyglot}}$ exploits this phenomenon to perform document~\cite{IndPak} and token~\cite{khudabukhsh2020harnessing}  level language identification with minimal supervision. Intuitively, the neighborhood of a word in a given language would be other words belonging to the same language, and  conducting naive nearest neighbor sampling is unlikely to reveal semantic relationships with words belonging to the other language. In fact, neither \emph{lucha} nor \emph{fight} features in each others top 1,000 naively sampled nearest neighbors.

\textbf{\emph{What happens when nearest neighbor searches are restricted to words in the corpus written in a different language than the input word?}} We discover that this simple \emph{constrained nearest neighbor sampling} is capable of retrieving substantial bilingual lexicons and this phenomenon is observed in multiple language pairs.

Formally, in a bilingual corpus $\mathcal{D}$ of two languages $\mathcal{L}_{\emph{source}}$ and $\mathcal{L}_{\emph{target}}$, let $\mathcal{V}_{\emph{source}}$ and $\mathcal{V}_{\emph{target}}$ denote the vocabularies of the two languages, respectively.  The translation scheme $\mathcal{L}_{\emph{source}} \rightarrow \mathcal{L}_{\emph{target}}$ takes a word $w_{\emph{source}}$ $\in$ $\mathcal{V}_{\emph{source}}$ as input and outputs a single word translation, $w_\emph{target}$, such that $w_\emph{target} \in \mathcal{V}_{\emph{target}}$ and $\forall w \in \mathcal{V}_{\emph{target}}$, \emph{dist}($w_{\emph{source}}$, $w$) $\ge$ \emph{dist}($w_{\emph{source}}$, $w_{\emph{target}}$). Following~\cite{mikolov2013efficient}, we take cosine distance as our distance metric (\emph{dist(.)}). Since we are operating on a single multilingual, noisy, social media corpus and Romanized Hindi does not have any standard spelling rules (e.g., the word \emph{amaan} meaning peace in Hindi can be spelled as \emph{aman} or \emph{amun}), we need a token-level language identification method to estimate the vocabularies of $\mathcal{L}_{\emph{source}}$ and $\mathcal{L}_{\emph{target}}$. For this, we use the minimally supervised  algorithm $\reallywidehat{\mathcal{L}}_{\emph{polyglot}}$. 

When we perform nearest neighbor sampling with this additional constraint, surprisingly, we find that the nearest neighbor of \emph{religion} is \emph{mazhab} and the nearest neighbor of \emph{lucha} is \emph{fight}! Note that, our method involves no explicit attempt to achieve alignment; a single Skip-gram model is trained on a bilingual corpus and lexicons are retrieved using this simple constrained nearest neighbor sampling. Further, noisy estimations of vocabularies are obtained using $\reallywidehat{\mathcal{L}}_{\emph{polyglot}}$.  Using our technique, we retrieve substantial bilingual lexicons across three different Indian social media corpora, and several (synthetically induced) bilingual corpora of European language pairs.  

\section{Notation}

We denote English, Spanish, German, Romanized Hindi as \emph{en}, \emph{es}, \emph{de}, and \emph{hi$_e$}, respectively. $\mathcal{L}_{\emph{source}}$ and $\mathcal{L}_{\emph{target}}$ indicate the direction of translation. For example, when translating $en \rightarrow es$,  $\mathcal{L}_{\emph{source}}$ is English and $\mathcal{L}_{\emph{target}}$ is Spanish. The vocabularies of the source and target languages are denoted by $\mathcal{V}_{\emph{source}}$ and $\mathcal{V}_{\emph{target}}$, respectively. The ground truth language label of a specific token $w$ or a document $d$ is returned by the function $\mathcal{L}$(.). When we use $\reallywidehat{\mathcal{L}}_{\emph{polyglot}}$ to estimate the language label or vocabulary, we indicate that by $\reallywidehat{\mathcal{L}}$(.) and $\reallywidehat{\mathcal{V}}$. In our work, we perform several frequency-based analyses to understand our method's strengths and weaknesses. We denote the top 5\% of words by frequency in the vocabulary as $\reallywidehat{\mathcal{V}}^{0-5}$ and similarly, the top 5-10\% by $\reallywidehat{\mathcal{V}}^{5-10}$.      
\section{Related Work}
Bilingual lexicon induction has a rich line of prior literature~\cite{rapp1999automatic,koehn2002learning,schafer-yarowsky-2002-inducing, fung-cheung-2004-mining,gaussier-etal-2004-geometric,haghighi-etal-2008-learning, vulic2013cross,vulic-etal-2011-identifying} with modern methods~\cite{mikolov2013exploiting,zou2013bilingual,faruqui2014improving,gouws2015bilbowa,xing-etal-2015-normalized,ammar2016massively,artetxe-etal-2017-learning,cao2016distribution,smith2017offline,conneau2017word,zhang-etal-2017-adversarial,joulin-etal-2018-loss,dou2018unsupervised} leveraging continuous representation of words~\cite{mikolov2013efficient,pennington2014glove,bojanowski2017enriching}. While this list is far from exhaustive, several of these focus on alleviating resource requirements like seed lexicon, or parallel data. For example,~\cite{artetxe-etal-2017-learning}  aligned two monolingual embeddings trained on distinct monolingual corpora using digits to address the requirement of large seed lexicons;~\cite{conneau2017word} aligned monolingual embeddings using a Cross-Domain Local Scaling measure that requires no seed lexicon, or parallel data. The resulting aligned embeddings perform favorably when compared against supervised methods. Our work contrasts with the literature in the following key ways. Unlike previous work, we embrace the challenge of learning bilingual lexicons from harsh, social media data. Our method does not require clean, monolingual corpora; we learn a single Skip-gram embedding on a bilingual corpus and present a novel finding that constrained nearest neighbor sampling can retrieve substantial bilingual lexicons without any explicit attempt to align. Our unsupervised method is particularly well-suited for noisy language expression typical of informal social media settings (e.g., Romanized Hindi) where procuring clean, monolingual (let alone parallel) data could be difficult. Recent work in unsupervised machine translation~\cite{lample-etal-2018-phrase} utilizes monolingual language models and alignment steps to learn a phrase-level translation model. Our work discards the individual monolingual models and the alignment steps instead using just a single polyglot Skip-gram model and a mining step. To reiterate, our motivations are not performance-driven - this paper explores the extent of multilingual information embedded in polyglot Skip-gram models.

As much as our paper is about bilingual lexicon induction using noisy, social media data, we also highlight the intriguing observation that polyglot embeddings can learn a rich semantic representation that captures word meanings across languages. While polyglot word-embeddings and polyglot training in particular
have received attention recently for demonstrating performance improvements across a variety of NLP tasks~\cite{mulcaire-etal-2019-polyglot, mulcaire-etal-2018-polyglot, mulcaire2019low,IndPak,khudabukhsh2020harnessing}, to the best of our knowledge, no previous work has explored their effectiveness in retrieving bilingual lexicons. 

Our work is related to~\cite{khudabukhsh2020harnessing} in two key ways: (1) use of $\reallywidehat{\mathcal{L}}_{\emph{polyglot}}$ to perform token-level language identification and (2) a shared task of cross-lingual sampling in the domain of \emph{hope speech}. However, unlike~\cite{khudabukhsh2020harnessing} who use vocabulary estimates to measure the extent of code mixing, we use the vocabulary estimates for our constrained nearest neighbor sampling. Moreover, our approach to perform cross-lingual sampling is different. We use our bilingual lexicons to construct noisy translations of English \emph{hope speech} while \cite{khudabukhsh2020harnessing} harness code switching. Finally, we obtain a 45\% performance improvement over the best-performing method reported by \cite{khudabukhsh2020harnessing}. 

\begin{table*}[htb]
{
\scriptsize
\begin{center}
     \begin{tabular}{|p{0.45\textwidth}|p{0.45\textwidth}|}
     \hline
     Sampled \emph{hope speech} & Loose translation \\
    \hline
\textbf{\texttt{\textcolor{red}{kaash dono mulko me dosti ho jaye dono mil kr \textcolor{black}{Europe} ke desho ki tra \textcolor{blue}{developed} ho skte hai \textcolor{blue}{piece love}}}}
 &     \emph{I wish two countries make friendship and together prosper and develop like European countries; peace, love.}\\
\hline
\textbf{\texttt{\textcolor{red}{kaash dono desho mein shanti k rishte kayam ho sake}}} & {\emph{I wish both countries can forge a relationship of peace.}} \\
\hline
\textbf{\texttt{\textcolor{red}{jung talti rhe to bahtar he ap or ham sabhi k anagan me shama jalti rhe to bahatr he jung to khud hi ek masla he jung kiya maslo ka hal de gi}}} & \emph{It is better if war is avoided. All of us should prosper. War in itself is a problem, how can it be a solution?}\\
    \hline
    \end{tabular}
    
\end{center}
\vspace{-0.1cm}
\caption{{Random sample of \emph{hope speech} obtained through our method.}}
\label{tab:hopeSpeech}}
\vspace{-0.25cm}
\end{table*}

\section{Data Sets}~\label{sec:data}
We use three Indian social media data sets. Two of them were introduced in prior literature. In addition, we construct a new data set.  


  \noindent\fooAlter \textbf{$\mathcal{D}_{\emph{hope}}$}  consists of 2.04 million comments posted by 791,289 user on 2,890 YouTube videos relevant to the 2019 India-Pakistan conflict~\cite{IndPak}. 

 \noindent\fooAlter \textbf{$\mathcal{D}_{\emph{election}}$}  consists of 6.18 million comments on 130,067 videos by 1,518,077 users posted in a  100 day period leading up to the 2019 Indian General Election~\cite{election}. 

\noindent\fooAlter \textbf{$\mathcal{D}_{\emph{covid}}$}  consists of 4,511,355 comments by 1,359,638 users on 71,969  videos from fourteen Indian news outlets (see, Appendix for details) posted between 30$^{\emph{th}}$ January, 2020\footnote{First COVID-19 positive case reported in India.} and 7$^{\emph{th}}$ May, 2020.

\subsection{Data set challenges}
As documented in~\cite{IndPak, election, khudabukhsh2020harnessing}, typical to most noisy, short social media texts generated in linguistically diverse regions, the data sets we consider exhibit a considerable presence of code mixing, and grammar and spelling disfluencies. On top of these, $\mathcal{D}_{\emph{hope}}$ and $\mathcal{D}_{\emph{election}}$ involve two additional challenges.  First, due to a strong presence of content contributors who do not speak English as their first language, varying levels of English proficiency in the corpus with a substantial incidence of phonetic spelling errors were reported. For example, 32\% of times, the word \texttt{liar} was misspelled as \texttt{lier}~\cite{election}, or consider the following example comment --  [\textbf{\texttt{\textcolor{blue}{pak pm godblashu my ind pailat thanksh you cantri}}}] that possibly intended to express \emph{Pak PM God bless you; my Ind pilot, thank you country}. We corroborated this finding on $\mathcal{D}_{\emph{covid}}$ where 31.8\% of the time \texttt{liar} was misspelled as \texttt{lier}.  Second, since Romanized Hindi does not have any standard spelling rules~\cite{gella2014ye} (e.g., the word \textcolor{red}{nuksaan} meaning \textcolor{blue}{damage} is spelled in the corpus as \textcolor{red}{nuksaan}, \textcolor{red}{nuqsaan} and \textcolor{red}{nuksan}), a high level of spelling variations added to the challenges. 
\subsection{Preprocessing}
For each of these three data sets as input, we conduct the following preprocessing and output the \emph{en} and \emph{hi}$_e$ vocabularies. 

\noindent\fooAlter We first clean the multilingual corpus with the same steps (e.g., removing emojis, lower casing words written in Roman script, removing punctuations) described in~\cite{election}.

\noindent\fooAlter Following~\cite{IndPak}, we train 100 dimensional FastText embeddings~\cite{bojanowski2017enriching} (full training configuration presented in Appendix). Following~\cite{khudabukhsh2020harnessing}, we construct the \emph{hi}$_e$ and \emph{en} vocabularies.


Note that, while our constrained nearest neighbor sampling restricts the vocabulary of the target word based on our obtained (noisy) vocabularies from $\reallywidehat{\mathcal{L}}_{\emph{polyglot}}$, we perform no explicit monolingual corpus extraction or removal of any other language expressed in Roman script or traditional script (e.g., Hindi in Devanagari constitutes a small fraction of all corpora, see Appendix for visualizations).

\section{Results}

\begin{table*}[htb]
\scriptsize
{

\begin{center}
     \begin{tabular}{|l | c  | c | c| c| c | c| }
     \hline 
     Measure & \emph{hi$_e$} $\rightarrow$\emph{en} & \emph{en} $\rightarrow$\emph{hi$_e$} &  \emph{hi$_e$}$\rightarrow$\emph{en} & \emph{en}$\rightarrow$\emph{hi$_e$} & \emph{hi$_e$}$\rightarrow$\emph{en} &
     \emph{en}$\rightarrow$\emph{hi$_e$}\\
    \hline
    &  $\mathcal{D}_{\emph{hope}}$ & $\mathcal{D}_{\emph{hope}}$ & $\mathcal{D}_{\emph{election}}$ & $\mathcal{D}_{\emph{election}}$ & $\mathcal{D}_{\emph{covid}}$ & $\mathcal{D}_{\emph{covid}}$  \\
     \hline 
  p@1 & 0.18 & 0.10 & 0.21 & 0.24 & 0.29 & 0.31  \\
   \hline
  p@5 & 0.39 &0.27  & 0.44 & 0.50 & 0.54 & 0.53 \\
   \hline 
  p@10 & 0.47 & 0.38& 0.52 & 0.61 & 0.63 & 0.63 \\
   \hline

    \end{tabular}

\end{center}
\caption{{Word translation performance on social media data. For each training corpus and translation direction, 500 source words are selected from $\protect\reallywidehat{{\mathcal{V}}}_{\emph{source}}^{0-5}$ and are mapped to target words in $\protect\reallywidehat{\mathcal{V}}_{\emph{target}}$ that are present in the corpus for at least 100 or more times. p@K indicates top-K accuracy.}}
\label{tab:socialMediaTranslation}}
\vspace{-0.4cm}
\end{table*}


\subsection{\emph{en}-\emph{hi$_e$} translation}

Table~\ref{tab:socialMediaTranslation} summarizes our performance in extracting bilingual lexicons across three Indian social media data sets. Following standard practice~\cite{smith2017offline, conneau2017word}, we report p@1, p@5 and p@10 performance. p@K is defined as the top-$K$ accuracy~\cite{mikolov2013exploiting}, i.e., an accurate translation of the source word is present in the retrieved top $K$ target words. It is common practice to restrict the vocabularies for source words (and target words) based on some prevalence criterion~\cite{conneau2017word}. We restrict $\reallywidehat{\mathcal{V}}_{source}$ to $\reallywidehat{\mathcal{V}}_{source}^{0-5}$ and $\reallywidehat{\mathcal{V}}_{target}$ to words that have appeared at least 100 times in the corpus (Appendix contains hyperparameter sensitivity analysis). 

As shown in Table~\ref{tab:socialMediaTranslation}, we observe that substantial bilingual lexicons can be retrieved using our unsupervised method. Even on multilingual, challenging social media data sets, a $p@5$ performance as high as 0.5 was achieved on multiple occasions. Across the three data sets, we obtain best performance in $\mathcal{D}_{\emph{covid}}$. Compared to other two data sets, $\mathcal{D}_{\emph{covid}}$ has stronger presence of \emph{hi}$_e$. Romanized Hindi does not have standard spelling rules; a larger volume of data could be useful in learning embeddings robust to spelling variations.

The three data sets we looked into have wildly different topical focus: international conflict, general election and a global pandemic. The nature of the successfully retrieved words also reflect this. From $\mathcal{D}_{\emph{hope}}$, we obtained translations for several conflict-words (e.g., \textcolor{blue}{attack}, \textcolor{blue}{war}, \textcolor{blue}{peace}, \textcolor{blue}{brave}, \textcolor{blue}{martyred}). In contrast, from  $\mathcal{D}_{\emph{election}}$ we obtained words focusing on national priorities and issues (e.g., \textcolor{blue}{corruption}, \textcolor{blue}{population}, \textcolor{blue}{unemployment}), while our method when applied on  $\mathcal{D}_{\emph{covid}}$ retrieved words focusing on disease symptoms, preventive measures and treatment terms (e.g., \textcolor{blue}{fever}, \textcolor{blue}{cough}, \textcolor{blue}{wash}, \textcolor{blue}{distance}, \textcolor{blue}{treatment}, \textcolor{blue}{medicine}). Our obtained lexicons had non-overlapping regions which indicate the possibility of growing the lexicon through combining multiple corpora focusing on topics with minimal overlap. We will release this bilingual lexicon (consensus label by two annotators, annotator details are present in the Appendix) of 1,100 unique word pairs as a resource. Table~\ref{tab:translatedPhrases} lists a few randomly chosen examples of  successfully translated word pairs across our three data sets.

\begin{table}[b]
\scriptsize
{
\begin{center}
     \begin{tabular}{| l  | l | l|}
     \hline 
     $\mathcal{D}_{\emph{hope}}$ & $\mathcal{D}_{\emph{election}}$ & $\mathcal{D}_{\emph{covid}}$\\
    \hline
     
  \textcolor{red}{aatankvadi} \textcolor{blue}{terrorist}   & \textcolor{red}{deshbhakti} \textcolor{blue}{patriotism} &
  \textcolor{red}{ilaz} \textcolor{blue}{treatment}
  
  	\\
   \hline
  \textcolor{red}{bahaduri} \textcolor{blue}{bravery} &
  \textcolor{red}{turant} \textcolor{blue}{immediately}
  & \textcolor{red}{joota} 
    \textcolor{blue}{shoe}
  	
  \\
   \hline
   \textcolor{red}{musalmano} \textcolor{blue}{muslims}    
   &\textcolor{red}{patrakaar}
   \textcolor{blue}{journalist} 	
   & \textcolor{red}{kahani}
   \textcolor{blue}{story}

   \\
   \hline
   \textcolor{red}{andha} \textcolor{blue}{blind}&
    \textcolor{red}{angrezo}	
   \textcolor{blue}{britishers}
  
& \textcolor{red}{jukam}	
   \textcolor{blue}{cold}
	
\\
   \hline
   \textcolor{red}{nuksan} \textcolor{blue}{damage}
   &\textcolor{red}{berojgari}
    \textcolor{blue}{unemployment}
   
   & 	
   \textcolor{red}{saf}
    \textcolor{blue}{clean}\\
   \hline
   \textcolor{red}{faida} \textcolor{blue}{benefit}
   &\textcolor{red}{ummeed}
    \textcolor{blue}{expectation}
   & 	 
   \textcolor{red}{hathon}
    \textcolor{blue}{hands}\\
    \hline
    \textcolor{red}{dino} 
    \textcolor{blue}{days} 
    &\textcolor{red}{nokri} 
    \textcolor{blue}{jobs} 	
&  	
\textcolor{red}{bachhe} 
    \textcolor{blue}{kids}\\
   \hline
   \textcolor{red}{bharosa} \textcolor{blue}{trust}
   &

    \textcolor{red}{bikash} \textcolor{blue}{devlopment}&
       \textcolor{red}{mudde} \textcolor{blue}{issues}\\
   \hline
    \textcolor{red}{tarakki} \textcolor{blue}{progress} 
    & \textcolor{red}{gareebi}	
      \textcolor{blue}{poverty}

    & 	
    \textcolor{red}{marij}	
      \textcolor{blue}{patient}\\
   \hline
   \textcolor{red}{gayab} \textcolor{blue}{vanish}
     & \textcolor{red}{shi} \textcolor{blue}{ryt}&
     \textcolor{red}{sankramit} \textcolor{blue}{infected}
     	\\
   \hline
\end{tabular}

\end{center}
\caption{{A random sample of word pairs translated by our algorithm. Appendix contains more examples.}}
\label{tab:translatedPhrases}}
\end{table}

\subsubsection{Qualitative Analysis}

In our translation scheme, we found that translations for nouns, adjectives and adverbs were successfully discovered (see Table~\ref{tab:translatedPhrases}). Preserving plurality (\textcolor{red}{hazaron} \textcolor{blue}{thousands}, \textcolor{red}{musalmano} \textcolor{blue}{muslims}, \textcolor{red}{naare} \textcolor{blue}{slogans}) on most occasions, translating numerals (\textcolor{red}{char} \textcolor{blue}{4}, \textcolor{red}{eik} \textcolor{blue}{one}) were among some surprising observations considering the noisy social media setting. 
For a given source word, multiple valid synonymous target words were often among the top translations produced by our method (e.g., \textcolor{red}{aman} and \textcolor{red}{shanti} for \textcolor{blue}{peace}; \textcolor{red}{dharam}, \textcolor{red}{mazhab} and  \textcolor{red}{jaat} for \textcolor{blue}{religion}). Stylistic choices like contraction were reflected in the translation (e.g., \textcolor{red}{kyki} (\emph{kyuki}) mapped to \textcolor{blue}{bcz} (\emph{because}), and \textcolor{red}{shi} (\emph{sahi}) mapped to \textcolor{blue}{ryt} (\emph{right})). Verbs are conjugated differently in Hindi and English and word-for-word translations don't typically exist - for instance \textcolor{blue}{help him} translates to \textcolor{red}{uska} \textcolor{blue}{``him''} \textcolor{red}{madad} \textcolor{blue}{``help''} \textcolor{red}{karo} \textcolor{blue}{``do''}, thus words like \textcolor{red}{karo} were rarely successfully translated. 


\noindent\textbf{Polysemy:} Our system performs single word translation. During translation, without context, detecting polysemy and resolving it to the true meaning w.r.t. the context is not possible. However, we were curious to learn if top translation choices of polysemous source words include valid translations of their different meanings. We notice that this was the case in a few instances. For example, the word \textcolor{blue}{cold} can mean both low temperature or a common viral infection. In $\mathcal{D}_{\emph{covid}}$, both these meanings were captured in the top translations while translations in $\mathcal{D}_{\emph{hope}}$ and $\mathcal{D}_{\emph{election}}$ only reflected the meaning of low temperature. It is unlikely that cold in the sense of viral infection would be highly discussed in the latter two corpora, while it is understandable that  common cold and associated symptoms would be heavily discussed in $\mathcal{D}_{\emph{covid}}$. 

\noindent\textbf{Nativization of loanwords:} Lexical borrowing across language pairs in the context of loanwords (or borrowed words) has been studied in linguistics~\cite{holden1976assimilation,calabrese2009loan,van2016loan} and computational linguistics~\cite{tsvetkov2016cross,patro-etal-2017-english}. Borrowed words, also known as loanwords, are lexical items borrowed from a donor language. For example, the English word \textcolor{blue}{avatar} or \textcolor{blue}{yoga} is borrowed from Hindi, while \textcolor{red}{botal} (\textcolor{blue}{bottle}) and \textcolor{red}{astabal} (\textcolor{blue}{stable}) are Hindi words borrowed from English. We noticed nativized loanwords, i.e., borrowed words that underwent  phonological repairs to adapt to a foreign language, translate back to their English donor counterpart (e.g., \textcolor{red}{rashan} and \textcolor{red}{angrezi} translate to donor words \textcolor{blue}{ration} and \textcolor{blue}{English}, respectively).           
\subsubsection{Digging deeper}

We conduct a detailed ablation study to understand this phenomenon. In what follows, we summarize our findings (see, Appendix for details).\\
\noindent\fooAlter \textbf{\emph{Disabling numbers}}: In prior literature,~\cite{artetxe-etal-2017-learning} showed that digits can be used as seed lexicons to align monolingual embeddings. Although our method doesn't make any explicit attempt to align, phrases like \textcolor{blue}{2019 election} (\textcolor{red}{2019 chunao}), \textcolor{blue}{1971 war} (\textcolor{red}{1971 jung}) can appear in both languages and hence can serve as signals. We replaced all numbers with a specific randomly chosen string that does not occur in the original corpus and evaluated the retrieval performance of our previously successful p@1 translations. We observed a performance dip of 28\% which indicates that numeric literals may contribute to this phenomenon.\\  
\noindent\fooAlter\textbf{\emph{Loanwords}}: We observe that in most cases, in successfully translated word pairs (e.g., $\langle \textcolor{red}{madad}, \textcolor{blue}{help} \rangle$, $\langle \textcolor{red}{ilaj}, \textcolor{blue}{treatment} \rangle$), at least one of the words is borrowed and used in the other language (e.g., {\textcolor{red}{humein}} {\textcolor{blue}{help}} {\textcolor{red}{chahiye}} ``we need help"). These loanwords thus result in similar contexts for word pairs from different languages - which are possibly reflected in the obtained word embeddings facilitating translation.\\
\noindent\fooAlter \textbf{\emph{Frequency preserving corpus transformation}}: We perform a frequency preserving loanword exchange (see, Appendix) to modify the corpus where translated word pairs are interchanged to diminish the extent of borrowing of these loanwords (e.g., phrases like {\textcolor{red}{humein}} {\textcolor{blue}{help}} {\textcolor{red}{chahiye}} is rewritten as {\textcolor{red}{humein}} {\textcolor{red}{madad}} {\textcolor{red}{chahiye}}). We observe that the p@1 performance dipped by 33\% after this corpus modification indicating that loanwords are possibly major contributors to this phenomenon.
\subsection{Cross-lingual sampling}
In our previous section, we have demonstrated that our unsupervised constrained nearest neighbor sampling method is able to retrieve substantial bilingual lexicons across three different \emph{en}-\emph{hi}$_{e}$ data sets. We now demonstrate a practical benefit of our method.  

We focus on the task of detecting hostility-diffusing \emph{hope speech} first introduced in~\cite{IndPak}. In a corpus  focusing on the 2019 India-Pakistan conflict, the authors advocated the importance of hostility-diffusing \emph{hope speech} and presented a \emph{hope speech} classifier for English content. While the authors present an important study of a modern conflict, much of the focus was centered around the English sub-corpus. In a recent study,~\cite{khudabukhsh2020harnessing} widened the analysis with a cross-lingual sampling technique to detect \emph{hope speech} content authored in Romanized Hindi. Using the English \emph{hope speech} classifier and leveraging $\reallywidehat{\mathcal{L}}_{\emph{polyglot}}$ to perform token-level language identification, their proposed method first detects highly code mixed \emph{hope speech}, then uses it to sample \emph{hi}$_e$ \emph{hope speech} using nearest-neighbor sampling (\emph{NN-sampling}) in the comment embedding space of the \emph{hi}$_e$ sub-corpus, $\mathcal{D}_{\emph{hope}}^{hi_e}$. 

\begin{algorithm}[h!]
\scriptsize
\DontPrintSemicolon
\SetAlgoLined
\textbf{Input:} A document $\mathcal{S}_{\emph{source}}$ denoted by [$w_1$,\ldots,$w_k$] \;
\textbf{Output:} A  document embedding of  $\mathcal{S}_{\emph{source}}$ translated into $\mathcal{L}_{\emph{target}}$ \;
\textbf{Dependency}: $\emph{topTranslations}(w_i, N)^{\mathcal{L}_{\emph{target}}}$ returns $N$ nearest neighbors of $\emph{embedding}(w_i)$ from $\reallywidehat{\mathcal{V}}_{\emph{target}}$ \;

\textbf{Initialization:} $\mathcal{E} \leftarrow \{\}$\;

\textbf{Main loop:}\;
\ForEach{\emph{word} ~w$_i$~$\in$~$\mathcal{S}_{{\textit{source}}}$ }
{
\eIf{$\hat{\mathcal{L}}(w_i)~=~\mathcal{L}_{\textit{target}}$}{$\mathcal{E} \gets \mathcal{E} \cup \{\emph{embedding}(w_i)\}$\;}
{$\mathcal{T} \gets \emph{topTranslations}(w_i, N)^{\mathcal{L}_{\emph{target}}}$\;
$\mathcal{C} \gets \{\}$\;
\ForEach{\emph{word} ~$w_t$~$\in$~$\mathcal{T}$ }
{
\If{$w_i~\in~\textit{topTranslations}(w_t, N)^{\mathcal{L}_{\emph{source}}}$}
{
$\mathcal{C} \gets \mathcal{C} \cup \{w_t\}$\;
}
}
\If{$\mathcal{C} \ne \{\}$}
{randomly select $w$ from $\mathcal{C}$\;
$\mathcal{E} \gets \mathcal{E} \cup \{\textit{embedding}(w)\}$\;
}
}
}
\textbf{Output:} Average of  $\mathcal{E}$\;
\caption{{ \small{\texttt{translateEmbedding}($\mathcal{S}_{\emph{source}}$)}}}
\label{algo:NN-sampling}
\end{algorithm}

\noindent\textbf{Baselines:} We include both methods proposed in~\cite{khudabukhsh2020harnessing} and their baseline (random sampling) for performance comparison.
\noindent\textbf{Our approach:} We take a set of English \emph{hope speech} comments, $\mathcal{A}$, as inputs and output a sample of Hindi \emph{hope speech} comments from the Romanized Hindi subset, $\mathcal{D}^{hi_e}_{\emph{hope}}$. For each English comment in $\mathcal{A}$, we compute a document embedding of a noisy translation in Hindi using our method - \texttt{translateEmbedding} described in Algorithm~\ref{algo:NN-sampling}. $N$ is set to 10. Once we obtain a set of translated embeddings, $\mathcal{B}$, similar to~\cite{khudabukhsh2020harnessing}, we perform \emph{NN-sampling} in the comment embedding space of $\mathcal{D}^{hi_e}_{\emph{hope}}$ (described in the Appendix). For fair comparison, in all our experiments, our parameter configurations are identical to those in~\cite{khudabukhsh2020harnessing}. 
\begin{table}[htb]
\scriptsize
{

\begin{center}
     \begin{tabular}{|l | c  |}
     \hline 
     Method & Performance  \\
    \hline
     
  \emph{random-Sample}($\mathcal{D}_{\emph{hope}}^{hi_e}$)$^\dagger$ & 1.8\%  \\
   \hline
   \emph{NN-Sample}($\mathcal{D}^{\emph{hope}}_{h_e, +}$)~\cite{khudabukhsh2020harnessing}$^\dagger$ & 21.88\%  \\
   \hline   
  \emph{NN-Sample}($\mathcal{D}^{\emph{hope}}_{+}$)~\cite{khudabukhsh2020harnessing}$^\dagger$ & 31.68\%  \\
\hline 
Only top translation choice without back-translation& 23.08\%\\
\hline 
Without back-translation & 25.36\%  \\
   \hline 
   Our method  & \textbf{46.04\%}  \\
   \hline

    \end{tabular}

\end{center}
\caption{{Sampling performance. Percentage of samples output by the algorithm that are judged correct by human. Results marked with symbol $\dagger$ are obtained from~\cite{khudabukhsh2020harnessing}.}}
\label{tab:samplingPerformance}}
\vspace{-0.3cm}
\end{table}

Our \texttt{translateEmbedding} algorithm is inspired by~\cite{smith2017offline} with some modifications. We now describe the intuitions behind our design choices. Performing a naive translation using our translation scheme runs into the well-studied hubness problem observed in high-dimensional spaces~\cite{dinu2014improving,jegou2008accurate, radovanovic2010hubs}. Essentially, the hubness problem arises when a small subset of words are ``universal'' neighbors and hence attract several many-to-one mappings. Existing strategies like mutual nearest neighbor~\cite{dinu2014improving} and a more involved method of Cross-Domain Similarity Local Scaling~\cite{conneau2017word} have been previously proposed to address this issue. We employ a simple mutual nearest neighbor technique, i.e., a source word's top translations should include only those target words that include the source word when translated back. For an input document, we obtain the noisy word-for-word translation and compute the resulting document's embedding - the average of the normalized word embeddings.

In our earlier analyses, we noticed that our translation scheme found several synonyms in its top choices (e.g., {\textcolor{blue}{religion}} has {\textcolor{red}{mazhab}}, {\textcolor{red}{dharam}} and {\textcolor{red}{jaat}} in its top translated choices). Also, since Romanized Hindi does not have standard spelling rules, the translations often contained prevalent spelling variations of the same word (e.g., {\textcolor{red}{aman}}, {\textcolor{red}{amaan}}, {\textcolor{red}{jung}}, {\textcolor{red}{jang}}). Moreover, our retrieved dictionaries are noisy with substantially better p@10 than p@1 performance. To account for this noise in translation and to induce more diversity, for each word, we randomly sample a mutually nearest neighbor.   

Table~\ref{tab:samplingPerformance} compares the performance of our sampling method against existing approaches. In order to shed light into influence of different design choices on performance, we present our main method and ablation studies after disabling random sampling from top choices and back-translation to address the hubness problem. We notice that strictly limiting ourselves to the top translation choice and not accounting for hubness yields substantially worse performance. Allowing more diversity through randomly selecting one of the top translation choices improves performance somewhat, however, the hubness problem appears to be the primary performance bottleneck. Our final algorithm as described in Algorithm~\ref{algo:NN-sampling} achieves the best performance. On a highly challenging task of mining rare positives (random sampling only yields 1.8\% \emph{hope speech}), we obtained a 45\% improvement over the previously reported best result.

\begin{figure}[t]
\centering
\includegraphics[trim={0 0 0 0},clip, height=1.25in]{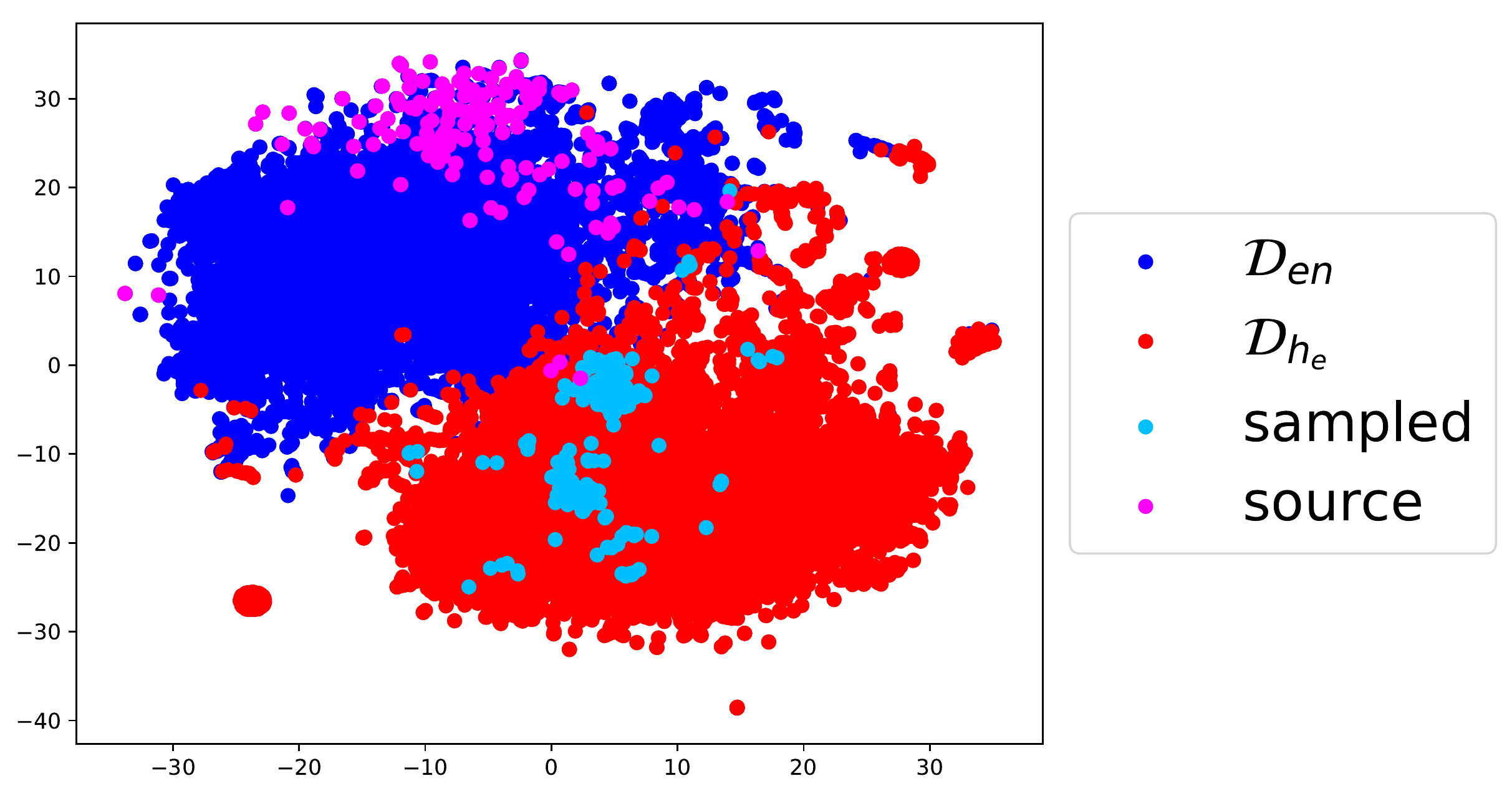}
\caption{{A 2D visualization showing the sampling results against the document embedding space.}}
\label{fig:cluster2}
\end{figure}

Table~\ref{tab:hopeSpeech} lists a random sample of \emph{hope speech} obtained using our method. We notice that the comments are mostly written in Romanized Hindi. A 2D visualization of the obtained comments indicate (see, Figure~\ref{fig:cluster2}) that our method retrieved comments reasonably distributed across the Hindi region.  


\subsection{Analyzing other language pairs}

We were curious to learn if our approach works with other language pairs. On two European language pairs, $\langle \emph{en}, \emph{es} \rangle$ and $\langle \emph{en}, \emph{de} \rangle$, we observed that our simple approach of constrained nearest neighbor sampling retrieves reasonable bilingual lexicons even when trained on a single, multilingual corpus (synthetically induced) without any explicit attempt to align. We acknowledge that prior art with supervision and seed lexicon (e.g., \cite{mikolov2013exploiting}) and recent unsupervised approaches (e.g., \cite{conneau2017word}) achieve considerably better performance on our test data set introduced in \cite{conneau2017word}. For instance, compared to our p@1 performance on \emph{en} $\rightarrow$ \emph{es} of 0.25,~\cite{mikolov2013exploiting} and ~\cite{conneau2017word} achieved performance of 0.33, and 0.82 respectively. Our focus is not on competing with the rich research conducted so far, rather, we are interested in reporting the generalizability of our approach. 

\noindent\textbf{Data sets:} We conduct experiments using Europarl~\cite{koehn2005europarl} and Wikipedia data sets. We synthetically induce a multilingual corpus by combining two monolingual corpora and then randomly shuffling at the sentence level. Table~\ref{tab:mainResult} summarizes our results. We find that our overall performance improved with Wikipedia data especially for \emph{de} $\rightarrow$ \emph{en} and \emph{es} $\rightarrow$ \emph{en}. \cite{conneau2017word} also reported a performance boost with Wikipedia data. 

\renewcommand{\tabcolsep}{1mm}
\begin{table}[t]
\scriptsize
{

\begin{center}
     \begin{tabular}{|l|l | c  | c | c| c| }
     \hline 
     Data set &  Measure & \emph{en} $\rightarrow $\emph{es} &  \emph{es} $\rightarrow$ \emph{en} &  \emph{en} $\rightarrow$ \emph{de} &
     \emph{de} $\rightarrow$ \emph{en}\\
     \hline 
  \multirow{3}{*}{Europarl}&p@1 & 0.25 & 0.25 & 0.19 & 0.16  \\ \cline{2-6}
  &p@5 & 0.37 & 0.39  & 0.30 & 0.18 \\\cline{2-6}
   
  &p@10 & 0.39 & 0.44 & 0.33 & 0.19 \\
   \hline 
   \hline
  \multirow{3}{*}{Wikipedia}&p@1 & 0.24 & 0.34 & 0.16 & 0.34  \\\cline{2-6}

  &p@5 & 0.40 & 0.50  & 0.31 & 0.46 \\\cline{2-6}
 
  &p@10 & 0.48 & 0.56 & 0.38 & 0.50 \\
   \hline

    \end{tabular}

\end{center}
\caption{{Performance comparison on Europarl~\cite{koehn2005europarl} and Wikipedia. $\protect\reallywidehat{\mathcal{V}}_{\emph{target}}$} is restricted to words that appeared more than 100 times in the corpus.}
\label{tab:mainResult}}
\vspace{-0.5cm}
\end{table}

In addition to an English-Italian translation retrieval task, we present an ablation  study in the Appendix. Our primary takeaways are:\\ 
\noindent\fooAlter 
\textbf{\emph{Source word frequency}}: Our experiments with Indian social media data sets indicate that our method performs better when we restrict ourselves to high-frequency source words. A fine-grained look at the performance based on the frequency of the source word reveals that we perform substantially better on high-frequency words belonging to  $\reallywidehat{\mathcal{V}}_{\emph{source}}^{0-5}$ (e.g., $en \rightarrow es$ performance jumps from 0.25 to 0.61 when we consider words in $\reallywidehat{\mathcal{V}}_{\emph{source}}^{0-5}$). 
\noindent\fooAlter \textbf{\emph{Topical cohesion}}: When we sample the \emph{en} part of the corpus from Europarl and the \emph{es} (or \emph{de}) part from Wikipedia, we remove the topical cohesion between the \emph{en} and \emph{es} (\emph{de}) components. We observe that performance dips slightly.


\section{Conclusions and Future Work}

In this paper, we explore the word embedding space resulting from training a single Skip-gram model on several bilingual corpora. Our detailed experiments reveal a rich and expressive embedding space across several language pairs that allows simple methods to retrieve substantial bilingual lexicons. In particular, the English-Romanized Hindi setting is a common occurrence in several corpora sourced in the Indian subcontinent. The (relatively) poorly resourced Romanized Hindi, and consequently the difficulty in obtaining monolingual corpora in this language pair make our observations and methods particularly well-suited for this setting. We explore intuitive but formidable cross-lingual sampling methods, and finally conduct detailed ablation studies for English-German and English-Spanish language pairs on existing data sets. We report that in all cases the rich embedding space is consistently observed and our methods just as applicable. Future directions include exploring the presence of this phenomenon in settings like contextual embeddings, and alternate models such as the highly successful transformer based methods.

\bibliographystyle{unsrt} 

\clearpage

\section{Appendix}

\begin{table*}[htb]
\scriptsize
{

\begin{center}
     \begin{tabular}{|l|l | c  | c | c| c| c | c| }
     \hline 
    \multirow{2}{*}{Measure}  &
    \multirow{2}{*}{\textbf{$\mathcal{V}_{\emph{source}}$}} & \emph{hi$_e$} $\rightarrow$\emph{en} & \emph{en} $\rightarrow$\emph{hi$_e$} &  \emph{hi$_e$}$\rightarrow$\emph{en} & \emph{en}$\rightarrow$\emph{hi$_e$} & \emph{hi$_e$}$\rightarrow$\emph{en} &
     \emph{en}$\rightarrow$\emph{hi$_e$}\\
    \cline{3-8}
    & & $\mathcal{D}_{\emph{hope}}$ & $\mathcal{D}_{\emph{hope}}$ & $\mathcal{D}_{\emph{election}}$ & $\mathcal{D}_{\emph{election}}$ & $\mathcal{D}_{\emph{covid}}$ & $\mathcal{D}_{\emph{covid}}$ \TBstrut \\
     \hline 
  \multirow{3}{*}{p@1} &\tiny{$\reallywidehat{\mathcal{V}}_{\emph{source}}^{0-5}$}& \cellcolor{blue!25}0.18 & \cellcolor{blue!25}0.10 & \cellcolor{blue!25}0.21 & 
  \cellcolor{blue!25}0.24 & 
  \cellcolor{blue!25}0.29 & 
  \cellcolor{blue!25}0.31 \TBstrut \\\cline{2-2}
   
   &\tiny{$\reallywidehat{\mathcal{V}}_{\emph{source}}^{5-10}$}& \cellcolor{red!25}0.11 & \cellcolor{red!25}0.02 & \cellcolor{red!25}0.22 & 
   \cellcolor{red!25}0.19 
   & \cellcolor{red!25}0.27 
   & \cellcolor{red!25}0.16 \TBstrut\\ \cline{2-2}
   &\tiny{$\reallywidehat{\mathcal{V}}_{\emph{source}}^{10-100}$}& \cellcolor{gray!25}0.09 & \cellcolor{gray!25}0.02 
   & \cellcolor{gray!25}0.06 & 
   \cellcolor{gray!25}0.11 & 
   \cellcolor{gray!25}0.07 & 
   \cellcolor{gray!25}0.05 \TBstrut\\ \hline
  \multirow{3}{*}{p@5} &\tiny{$\reallywidehat{\mathcal{V}}_{\emph{source}}^{0-5}$}& \cellcolor{blue!25}0.39 &\cellcolor{blue!25}0.27  & \cellcolor{blue!25}0.44 & \cellcolor{blue!25}0.50 & \cellcolor{blue!25}0.54 & \cellcolor{blue!25}0.53 \TBstrut\\\cline{2-2}
  &\tiny{$\reallywidehat{\mathcal{V}}_{\emph{source}}^{5-10}$}& \cellcolor{red!25}0.26& \cellcolor{red!25}0.15  & \cellcolor{red!25}0.43 & \cellcolor{red!25}0.33 & \cellcolor{red!25}0.45 & \cellcolor{red!25}0.43 \TBstrut\\\cline{2-2}
  &\tiny{$\reallywidehat{\mathcal{V}}_{\emph{source}}^{10-100}$}& \cellcolor{gray!25}0.20 & \cellcolor{gray!25}0.07  & \cellcolor{gray!25}0.18 & \cellcolor{gray!25}0.21 & \cellcolor{gray!25}0.23 & \cellcolor{gray!25}0.11 \TBstrut\\
   \hline 
  \multirow{3}{*}{p@10} &\tiny{$\reallywidehat{\mathcal{V}}_{\emph{source}}^{0-5}$}& \cellcolor{blue!25}0.47 & \cellcolor{blue!25}0.38& \cellcolor{blue!25}0.52 & \cellcolor{blue!25}0.61 & \cellcolor{blue!25}0.63 & \cellcolor{blue!25}0.63 \TBstrut\\\cline{2-2}
  &\tiny{$\reallywidehat{\mathcal{V}}_{\emph{source}}^{5-10}$}&\cellcolor{red!25}0.35 & \cellcolor{red!25}0.20& \cellcolor{red!25}0.48& \cellcolor{red!25}0.46 & \cellcolor{red!25}0.50 & \cellcolor{red!25}0.53 \TBstrut\\\cline{2-2}
  &\tiny{$\reallywidehat{\mathcal{V}}^{10-100}$}& \cellcolor{gray!25}0.22 & \cellcolor{gray!25}0.10 & \cellcolor{gray!25}0.22 & \cellcolor{gray!25}0.26 & \cellcolor{gray!25}0.26 & \cellcolor{gray!25}0.13 \TBstrut\\
   \hline

    \end{tabular}

\end{center}
\caption{{Word translation performance on social media data. Each cell summarizes the $p@K$ performance for a given translation direction on a data set as $a$ / $b$ / $c$, where $a$
(top) is the performance observed when the source vocabulary is restricted to $\protect\reallywidehat{\mathcal{V}}_{\emph{source}}^{0-5}$  (color coded with blue); $b$ (middle) is the performance observed when the source vocabulary is restricted to $\protect\reallywidehat{\mathcal{V}}_{\emph{source}}^{5-10}$ (color coded with red); c (bottom) is the performance observed when the source vocabulary is restricted to $\protect\reallywidehat{\mathcal{V}}_{\emph{source}}^{10-100}$(color coded with gray). 500 source words are randomly selected from  $\protect\reallywidehat{\mathcal{V}}_{\emph{source}}^{0-5}$; from $\protect\reallywidehat{\mathcal{V}}_{\emph{source}}^{5-10}$ and  $\protect\reallywidehat{\mathcal{V}}_{\emph{source}}^{10-100}$, 100 source words are randomly selected. The selected words are mapped to target words in $\protect\reallywidehat{\mathcal{V}}_{\emph{target}}$ that are present in the corpus for at least 100 or more times. p@K indicates top-K accuracy.}}
\label{tab:socialMediaTranslationFineGrained}}
\end{table*}

\renewcommand{\tabcolsep}{1mm}
\begin{table}[h!]
\scriptsize
{

\begin{center}
     \begin{tabular}{|l|l | c  | c | c| c| }
     \hline 
     Data set &  Measure & \emph{en} $\rightarrow $\emph{es} &  \emph{es} $\rightarrow$ \emph{en} &  \emph{en} $\rightarrow$ \emph{de} &
     \emph{de} $\rightarrow$ \emph{en}\\
     \hline \Tstrut
  \multirow{12}{*}{Europarl}& \multirow{4}{*}{p@1} & 0.25 & 0.25 & 0.19 & 0.16  \\ 
  & & \cellcolor{blue!25}0.61 & \cellcolor{blue!25}0.26  & \cellcolor{blue!25}0.43 & \cellcolor{blue!25}0.15 \\
   
  & & \cellcolor{red!25}0.50 & \cellcolor{red!25}0.28 & \cellcolor{red!25}0.39 & \cellcolor{red!25}0.12 \\
  & & \cellcolor{gray!25}0.17 & \cellcolor{gray!25}0.24 & \cellcolor{gray!25}0.13 & \cellcolor{gray!25}0.22 \\\cline{2-6}
  \Tstrut
& \multirow{4}{*}{p@5} & 0.37 & 0.39 & 0.30 & 0.18  \\ 
  & & \cellcolor{blue!25}0.79 & \cellcolor{blue!25}0.58  & \cellcolor{blue!25}0.68 & \cellcolor{blue!25}0.38 \\
   
  & & \cellcolor{red!25}0.70 & \cellcolor{red!25}0.52 & \cellcolor{red!25}0.65 & \cellcolor{red!25}0.34 \\
  & & \cellcolor{gray!25}0.26 & \cellcolor{gray!25}0.33 & \cellcolor{gray!25}0.19 & \cellcolor{gray!25}0.28 \\\cline{2-6}
  &  \Tstrut \multirow{4}{*}{p@10} & 0.39 &0.44 &0.33 &0.19  \\ 
  & & \cellcolor{blue!25}0.83 & \cellcolor{blue!25}0.66  & \cellcolor{blue!25}0.73 & \cellcolor{blue!25}0.50 \\
   
  & & \cellcolor{red!25}0.76 & \cellcolor{red!25}0.59 & 0.70 \cellcolor{red!25}& \cellcolor{red!25}0.43 \\
  & & \cellcolor{gray!25}0.28 & \cellcolor{gray!25}0.37 & \cellcolor{gray!25}0.22 & \cellcolor{gray!25}0.31 \\\cline{2-6}
  \hline 
  
    \end{tabular}

\end{center}
\caption{{Performance summary of our approach with training data set Europarl~\cite{koehn2005europarl}; test data set (denoted by $\protect\mathcal{D}_{\emph{test}}$) introduced in~\cite{conneau2017word}. $\protect\reallywidehat{\mathcal{V}}_{\emph{target}}$} is restricted to words that appeared more than 100 times in the training data set. Each cell summarizes the $p@K$ performance  for a given translation direction as $a$ / $b$ / $c$ / $d$, where $a$ (top)
is the overall performance observed on $\protect\mathcal{D}_{\emph{test}}$; $b$
is the performance observed on $\protect\reallywidehat{\mathcal{V}}_{\emph{source}}^{0-5} \cap \protect\mathcal{D}_{\emph{test}}$ (color coded with blue); $c$
is the performance observed on $\protect\reallywidehat{\mathcal{V}}_{\emph{source}}^{5-10} \cap \protect\mathcal{D}_{\emph{test}}$ (color coded with red); d (bottom) is the performance observed on $\protect\reallywidehat{\mathcal{V}}_{\emph{source}}^{10-100} \cap \protect\mathcal{D}_{\emph{test}}$ (color coded with gray).  }
\label{tab:finegrainedEuropean}}
\end{table}

\subsection{$\mathcal{D}_{\emph{covid}}$ data set}

We used the publicly available YouTube API to crawl our comments. We focused on the following date range: 30$^{\emph{th}}$ January, 2020\footnote{First COVID-19 positive case reported in India.} and 7$^{\emph{th}}$ May, 2020. Our data set consists of 4,511,355 comments by 1,359,638 users on 71,969  videos from fourteen Indian news outlets listed in Table~\ref{tab:nationalChannels}.

\begin{figure}[htb]
\centering
\includegraphics[trim={0 0 0 0},clip, height=1.25in]{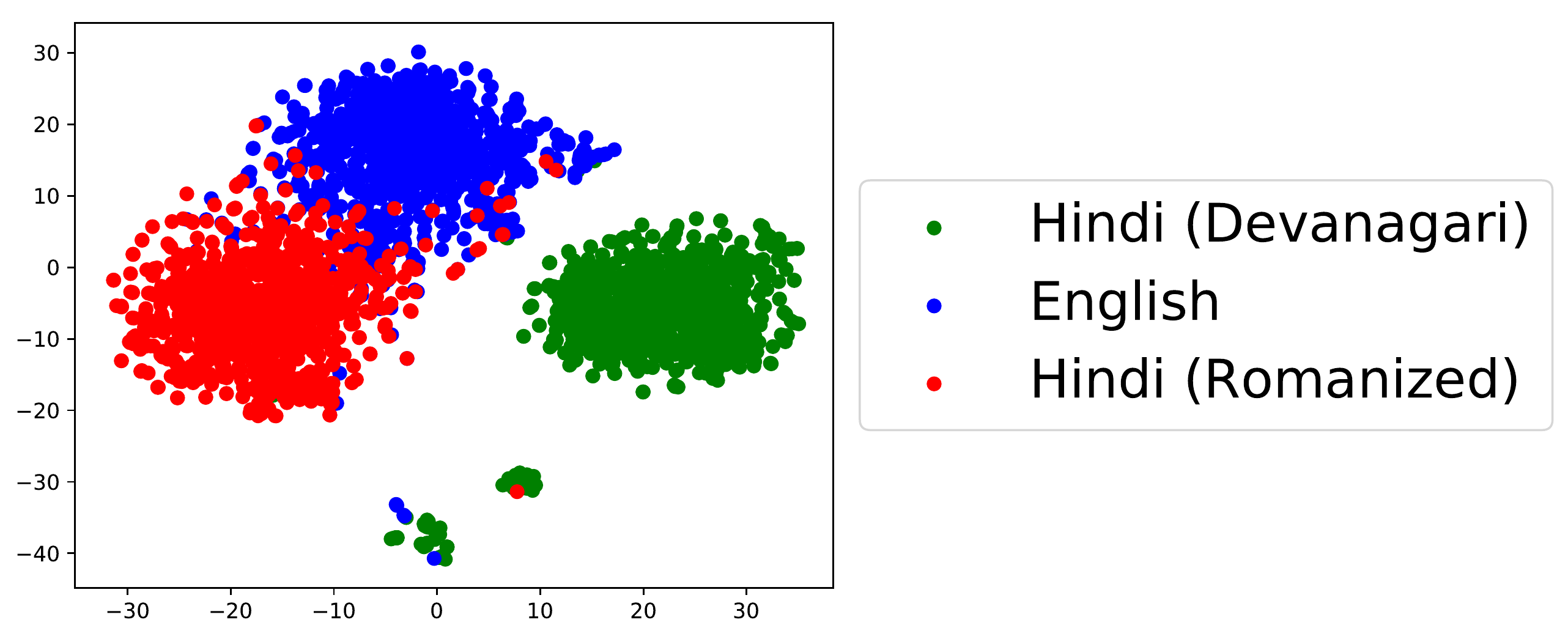}
\caption{{A 2D visualization of $\mathcal{D}_{\emph{covid}}$. Apart from English and Romanized Hindi, Hindi in Devanagari also has substantial presence in the corpus.}}
\label{fig:covidCluster}
\vspace{-0.5cm}
\end{figure}

\begin{table}[htb]

{
\begin{center}
     \begin{tabular}{|l | l  | l| l|}
     \hline 
     Corpus & $\reallywidehat{\mathcal{V}}$ for \emph{en} & $\reallywidehat{\mathcal{V}}$ for \emph{hi}$_e$ & $\reallywidehat{\mathcal{V}}$ for \emph{hi}$_e$ \\
     \hline 
      $\mathcal{D}_{\emph{hope}}$ & 38,516 & 71,677 & 23,560\\
      \hline 
     $\mathcal{D}_{\emph{election}}$ & 55,164 & 109,341 & 45,467\\
      \hline 
     $\mathcal{D}_{\emph{covid}}$ &46,504 & 109,809 & 59,219 \\
      \hline 
    \end{tabular}

\end{center}
\caption{Size of the estimated vocabularies using $\protect\reallywidehat{\mathcal{L}}_{\emph{polyglot}}$ on our data sets. Spelling variations in Romanized Hindi possibly contributed to a large size of Romanized Hindi vocabulary.}
\label{tab:vocabStats}}
\end{table}

Figure~\ref{fig:covidCluster} presents a 2D visualization of the word embeddings obtained using $\reallywidehat{\mathcal{L}}_{\emph{polyglot}}$. The visualization indicates that apart from Romanized Hindi and English, our data set also demonstrates substantial presence of Hindi written in Devanagari script further establishing the challenges associated to our task. The size of the estimated vocabularies is presented in Table~\ref{tab:vocabStats}. 

\begin{table}[h!]

{
\scriptsize
\begin{center}
     \begin{tabular}{|p{7cm}| }
     \hline 
    IndiaTV,
   NDTV India,
  Republic World,
  The Times of India,
  Zee News,
Aaj Tak,
ABP NEWS,
CNN-News18,
News18 India,
NDTV,
TIMES NOW,
India Today,
The Economic Times,
Hindustan Times \\ 
\hline 

    \end{tabular}

\end{center}
\caption{National channels.}
\label{tab:nationalChannels}}
\end{table}

\subsection{Embedding Hyperparameters}

All the models discussed in this paper are obtained by training Fasttext Skip-gram models with the following parameters unless stated otherwise:

\begin{compactitem}
    \item Dimension: 100
    \item Minimum subword unit length: 2
    \item Maximum subword unit length: 4
    \item Epochs: 5
    \item Context window: 5
    \item Number of negatives sampled: 5
\end{compactitem}


\subsection{Hyperparameter sensitivity analysis}

Recall that, we restricted $\reallywidehat{\mathcal{V}}_{\emph{source}}$ and $\reallywidehat{\mathcal{V}}_{\emph{target}}$ to prevalence criteria that (1) $\reallywidehat{\mathcal{V}}_{\emph{source}}$ is restricted to $\reallywidehat{\mathcal{V}}_{\emph{source}}^{0-5}$ (2) $\reallywidehat{\mathcal{V}}_{\emph{target}}$ contains words that have appeared at least 100 or more times in the corpus. In Table~\ref{tab:socialMediaTranslationFineGrained}, we relax the prevalence criterion on $\reallywidehat{\mathcal{V}}_{\emph{source}}$ and observe that as we move towards more infrequent words, the translation performance degrades. The performance drop is more visible with $\reallywidehat{\mathcal{V}}_{\emph{source}}^{10-100}$. Our annotators informed that poor quality of spelling and increased prevalence of contraction made the annotation task particularly challenging for rare words. 

We next analyze the effect of the frequency threshold of 100 on $\reallywidehat{\mathcal{V}}_{\emph{target}}$. In order to reduce annotation burden, we only focused on the subset of words with perfect translation (i.e., p@1 performance 100\%). When we relax the frequency threshold to 10, our p@1, p@5 and p@10 numbers are respectively, 0.38, 0.84, 0.91, respectively. Hence, although for 91\% or the source words we found a translation within the top 10 translations, our p@1 performance took a considerable hit.  Our annotators reported that with a lowered frequency threshold, the retrieved translations contained higher degree of misspellings. Our conclusion from this experiment is 100 is a reasonable threshold given the noisy nature of our corpora.

We conducted a similar analysis on our word translation tasks using European language pairs. As shown in Table~\ref{tab:finegrainedEuropean}, when English is the source language, our translation performance on frequent words is substantially better than rare words. However, when English is the target language, we did not observe any similar trend, the performance was roughly equal across the entire spectrum of words ranked by frequency. With Wikipedia corpus (not shown in the Table), we observed qualitatively similar trends.

\subsection{Annotation}

We used two annotators fluent in Hindi, Urdu and English. For word translations, consensus labels were used. For \emph{hope speech} annotation, the minimum Fleiss' $\kappa$ measure was high (0.88) indicating strong inter-rater agreement. After independent labeling, differences were resolved through discussion.

\subsection{Extended examples of lexicons}

Table~\ref{tab:translatedPhrasesExtended} lists an extended bilingual lexicon containing 90 words pairs (30 from each corpus) obtained using our method. We will release the complete lexicon of 1,100 word pairs upon acceptance.

\begin{table}[htb]
\scriptsize
{
\begin{center}
     \begin{tabular}{| l  | l | l|}
     \hline 
     $\mathcal{D}_{\emph{hope}}$ & $\mathcal{D}_{\emph{election}}$ & $\mathcal{D}_{\emph{covid}}$\\
    \hline
     
  \textcolor{red}{aatankvadi} \textcolor{blue}{terrorist}   & \textcolor{red}{deshbhakti} \textcolor{blue}{patriotism} &
  \textcolor{red}{ilaz} \textcolor{blue}{treatment}
  
  	\\
   \hline
  \textcolor{red}{bahaduri} \textcolor{blue}{bravery} &
  \textcolor{red}{turant} \textcolor{blue}{immediately}
  & \textcolor{red}{joota} 
    \textcolor{blue}{shoe}
  	
  \\
   \hline
   \textcolor{red}{musalmano} \textcolor{blue}{muslims}    
   &\textcolor{red}{patrakaar}
   \textcolor{blue}{journalist} 	
   & \textcolor{red}{kahani}
   \textcolor{blue}{story}

   \\
   \hline
   \textcolor{red}{andha} \textcolor{blue}{blind}&
    \textcolor{red}{angrezo}	
   \textcolor{blue}{britishers}
  
& \textcolor{red}{jukam}	
   \textcolor{blue}{cold}
	
\\
   \hline
   \textcolor{red}{nuksan} \textcolor{blue}{damage}
   &\textcolor{red}{berojgari}
    \textcolor{blue}{unemployment}
   
   & 	
   \textcolor{red}{saf}
    \textcolor{blue}{clean}\\
   \hline
   \textcolor{red}{faida} \textcolor{blue}{benefit}
   &\textcolor{red}{ummeed}
    \textcolor{blue}{expectation}
   & 	 
   \textcolor{red}{hathon}
    \textcolor{blue}{hands}\\
    \hline
    \textcolor{red}{dino} 
    \textcolor{blue}{days} 
    &\textcolor{red}{nokri} 
    \textcolor{blue}{jobs} 	
&  	
\textcolor{red}{bachhe} 
    \textcolor{blue}{kids}\\
   \hline
   \textcolor{red}{bharosa} \textcolor{blue}{trust}
   &

    \textcolor{red}{bikash} \textcolor{blue}{devlopment}&
       \textcolor{red}{mudde} \textcolor{blue}{issues}\\
   \hline
    \textcolor{red}{tarakki} \textcolor{blue}{progress} 
    & \textcolor{red}{gareebi}	
      \textcolor{blue}{poverty}

    & 	
    \textcolor{red}{marij}	
      \textcolor{blue}{patient}\\
   \hline
   \textcolor{red}{gayab} \textcolor{blue}{vanish}
     & \textcolor{red}{shi} \textcolor{blue}{ryt}&
     \textcolor{red}{sankramit} \textcolor{blue}{infected}
     	\\
   \hline
   \textcolor{red}{kyki} \textcolor{blue}{bcz}
     & \textcolor{red}{bazar} \textcolor{blue}{market}&
     \textcolor{red}{hoshiyar} \textcolor{blue}{smart}
     	\\
   \hline

   \textcolor{red}{jahannam} \textcolor{blue}{hell}
     & \textcolor{red}{masoom} \textcolor{blue}{innocent}&
     \textcolor{red}{khubsurat} \textcolor{blue}{beautiful}
     	\\

\hline
   \textcolor{red}{tel} \textcolor{blue}{oil}
     & \textcolor{red}{sanghatan} \textcolor{blue}{organization}&
     \textcolor{red}{dange} \textcolor{blue}{riots}
     	\\

\hline
   \textcolor{red}{darr} \textcolor{blue}{fear}
     & \textcolor{red}{chhavi} \textcolor{blue}{image}&
     \textcolor{red}{bilkul} \textcolor{blue}{absolutely}
     	\\

\hline
   \textcolor{red}{halat} \textcolor{blue}{condition}
     & \textcolor{red}{mahina} \textcolor{blue}{month}&
     \textcolor{red}{arakshan} \textcolor{blue}{reservation}
     	\\

\hline
   \textcolor{red}{intzaar} \textcolor{blue}{wait}
     & \textcolor{red}{qatal} \textcolor{blue}{murder}&
     \textcolor{red}{palan} \textcolor{blue}{obey}
     	\\

\hline
   \textcolor{red}{sipahi} \textcolor{blue}{soldier}
     & \textcolor{red}{hinsak} \textcolor{blue}{violent}&
     \textcolor{red}{maut} \textcolor{blue}{death}
     	\\

\hline
   \textcolor{red}{peety} \textcolor{blue}{drinking}
     & \textcolor{red}{bohot} \textcolor{blue}{very}&
     \textcolor{red}{sadasya} \textcolor{blue}{member}
     	\\

\hline
   \textcolor{red}{gau} \textcolor{blue}{cows}
     & \textcolor{red}{garibon} \textcolor{blue}{poors}&
     \textcolor{red}{achanak} \textcolor{blue}{suddenly}
     	\\

\hline
   \textcolor{red}{jawab} \textcolor{blue}{answer}
     & \textcolor{red}{blot} \textcolor{blue}{dhabba}&
     \textcolor{red}{dost} \textcolor{blue}{friend}
     	\\

\hline
   \textcolor{red}{alag} \textcolor{blue}{separate}
     & \textcolor{red}{chokidaar} \textcolor{blue}{watchman}&
     \textcolor{red}{hinsa} \textcolor{blue}{violence}
     	\\

\hline
   \textcolor{red}{pahele} \textcolor{blue}{first}
     & \textcolor{red}{shabd} \textcolor{blue}{word}&
     \textcolor{red}{behad} \textcolor{blue}{extremely}
     	\\

\hline
   \textcolor{red}{farq} \textcolor{blue}{difference}
     & \textcolor{red}{fela} \textcolor{blue}{spread}&
     \textcolor{red}{bukhar} \textcolor{blue}{fever}
     	\\

\hline
   \textcolor{red}{banana} \textcolor{blue}{make}
     & \textcolor{red}{peshaab} \textcolor{blue}{urine}&
     \textcolor{red}{bhedbhav} \textcolor{blue}{discrimination}
     	\\

\hline
   \textcolor{red}{sahi} \textcolor{blue}{right}
     & \textcolor{red}{niyam} \textcolor{blue}{regulations}&
     \textcolor{red}{vakeel} \textcolor{blue}{lawyer}
     	\\

\hline
   \textcolor{red}{panah} \textcolor{blue}{shelter}
     & \textcolor{red}{mouka} \textcolor{blue}{chance}&
     \textcolor{red}{taqat} \textcolor{blue}{strength}
     	\\

\hline
   \textcolor{red}{khao} \textcolor{blue}{eat}
     & \textcolor{red}{pehla} \textcolor{blue}{1st}&
     \textcolor{red}{aurat} \textcolor{blue}{woman}
     	\\

\hline
   \textcolor{red}{sadak} \textcolor{blue}{road}
     & \textcolor{red}{bahas} \textcolor{blue}{debate}&
     \textcolor{red}{unpadh} \textcolor{blue}{uneducated}
     	\\

\hline
   \textcolor{red}{shukar} \textcolor{blue}{thanks}
     & \textcolor{red}{akhri} \textcolor{blue}{last}&
     \textcolor{red}{sabkuch} \textcolor{blue}{everything}
     	\\

\hline
   \textcolor{red}{dhyan} \textcolor{blue}{focus}
     & \textcolor{red}{gotala} \textcolor{blue}{scam}&
     \textcolor{red}{sanskriti} \textcolor{blue}{culture}
     	\\

   \hline

\end{tabular}

\end{center}
\caption{{A random sample of translated word pairs from our corpora.}}
\label{tab:translatedPhrasesExtended}}
\end{table}

\subsection{Loanword}

We now slightly abuse the definition of a loanword and consider a word is a loanword if it appears in a context of words written in a different language, and define a simple measure to quantify to what extent this occurs in a two-language setting. Let  $c$  denote the  context (single word left and right) of a word $w$. We first count the  instances where the language labels of $c$ and $w$ agree, i.e., $\mathcal{L}({w}) = \mathcal{L}({c})$ (e.g., \textcolor{blue}{help} is not a loanword in the following phrase: {\textcolor{blue}{please help us}}). Let this number be denoted as $\mathcal{N}_{\emph{not-borrowed}}$. Similarly, we count the instances when $c$ and $w$ have different language labels, i.e.,  $\mathcal{L}({w}) \ne \mathcal{L}({c})$. This scenario would arise when a word is borrowed from a different language (e.g., \textcolor{blue}{help} is a loanword in {\textcolor{red}{humein}} {\textcolor{blue}{help}} {\textcolor{red}{chahiye}}). In our scheme, the Loan Word Index (\emph{LWI}) of a word $w$ is defined as \emph{LWI}($w$) = $\frac{\mathcal{N}_{\emph{borrowed}}}{\mathcal{N}_{\emph{borrowed}} + \mathcal{N}_{\emph{not-borrowed}}}$. A high \emph{LWI} indicates substantial lexical borrowing of the word outside its language. Since we use  $\reallywidehat{\mathcal{L}}_{\emph{polyglot}}$ to estimate language labels, we indicate  \emph{LWI}(.) as $\reallywidehat{\emph{LWI}}(.)$. For a word pair $\langle w_{\emph{source}}, w_{\emph{target}}  \rangle$, we define \emph{LWI}(.) as the maximum of their individual \emph{LWI}s. For example, if the \emph{LWI} is high for the pair $\langle \textcolor{blue}{help}, \textcolor{red}{madad} \rangle$, it indicates that at least one of these words was substantially borrowed. Our hypothesis is that successfully translated word pairs would have a high \emph{LWI} indicating at least one of the pair was used as a loanword facilitating translation. The average Loan Word Index of all successfully translated word pairs in our test data sets across all three corpora is 0.15. Compared to this, randomly generated word pairings have an average Loan Word Index of 0.09. We next performed a frequency preserving loan word exchange to modify the corpus where translated word pairs are interchanged to diminish the extent to which words are borrowed (e.g., phrases like {\textcolor{red}{humein}} {\textcolor{blue}{help}} {\textcolor{red}{chahiye}} is rewritten as {\textcolor{red}{humein}} {\textcolor{red}{madad}} {\textcolor{red}{chahiye}}). Frequency is preserved by interchanging both words in a successfully translated word pair as many times as the least borrowed word is borrowed. In our example if {\textcolor{red}{madad}} was borrowed 10 times, and {\textcolor{blue}{help}} 15 times, we alter 10 instances where {\textcolor{red}{madad}} is borrowed with {\textcolor{blue}{help}}, and 10 instances where {\textcolor{blue}{help}} is borrowed with {\textcolor{red}{madad}}. We thus preserve word frequencies while diminishing the loanword phenomenon. We observed that the retrieval performance of our p@1 set dipped by 33\% after this corpus modification indicating that frequent borrowing of words possibly positively contributed to our method's translation performance.

\subsection{\emph{NN-sample}}

\begin{algorithm}[tb]

\DontPrintSemicolon
\SetAlgoLined

\textbf{Initialization:}\; 
$\mathcal{E} \leftarrow \{\}$\;

\textbf{Main loop:}\;
\ForEach{\emph{comment} ~c~$\in$~$\mathcal{S}$ }
{
$count \leftarrow 0$\;
$dist \leftarrow 0$\;
\While{count $\leq$ size}
{
    $neighbor \gets getNearestNeighbor(c, dist)$\;
    $dist \gets cosineDistance(c, neighbor)$\;
    
    \If{neighbor $\notin$ $\mathcal{E} \cup \mathcal{S}$}{$\mathcal{E} \gets \mathcal{E} \cup \{neighbor\}$\;
    $count \gets count + 1$\;}
  }
}
\textbf{Output:} $\mathcal{E}$
\caption{{ {\texttt{NN-Sample}($\mathcal{S}$, $\mathcal{U}$)}}}

\label{algo:NN-sampling-algo}

\end{algorithm}

Algorithm~\ref{algo:NN-sampling-algo} reproduces the \emph{NN-Sample} (we refer to this as \emph{NN-Sampling} in our paper) method as presented in~\cite{khudabukhsh2020harnessing}.
The method takes a seed set of documents $\mathcal{S}$ and a pool of documents $\mathcal{U}$ as inputs, and outputs $\mathcal{E}$ $\subset$ $\mathcal{U}$, containing nearest neighbors of $\mathcal{S}$ in the comment-embedding space. Initially, $\mathcal{E}$ is an empty set and at each step, $\mathcal{E}$ is expanded with nearest neighbors that are not present in the expanded set or the seed set. Consistent with ~\cite{khudabukhsh2020harnessing}, the parameter $\emph{size}$ is set to 5, and $\mathcal{U}$ is set to $\mathcal{D}_{hope}^{{hi}_e}$. 


\subsection{Topical cohesion}

We break topical cohesion by sampling \emph{en} and \emph{es} (\emph{de}) from Europarl and Wikipedia respectively. Our results show that bilingual lexicons are still retrieved albeit with marginally lower performance. We conclude that topical cohesion possibly helps but may not be a prerequisite for retrieving a reasonably sized bilingual lexicon.

\renewcommand{\tabcolsep}{2mm}
\begin{table*}[htb]
\scriptsize
{

\begin{center}
     \begin{tabular}{|l | c  | c | c| c| c | c| c| c|c| c|c|c|}
     \hline 
     Measure & \multicolumn{3}{c|}{\emph{en} $\rightarrow$\emph{es}}   & \multicolumn{3}{c|}{\emph{es} $\rightarrow$\emph{en}}   &  \multicolumn{3}{c|}{\emph{en} $\rightarrow$\emph{de}}
     & \multicolumn{3}{c|}{\emph{de} $\rightarrow$\emph{en}}\\
    \hline
    p@1& \cellcolor{blue!25}0.25 & \cellcolor{red!25}0.24 & \cellcolor{gray!25}0.14 & 
    \cellcolor{blue!25}0.25 & 
    \cellcolor{red!25}0.34 & \cellcolor{gray!25}0.21 
     & \cellcolor{blue!25}0.19 & \cellcolor{red!25}0.16 & \cellcolor{gray!25}0.10
     &\cellcolor{blue!25}0.16 & \cellcolor{red!25}0.34 & \cellcolor{gray!25}0.27
\\   
\hline
p@5 & 
\cellcolor{blue!25}0.37 & \cellcolor{red!25}0.40 & \cellcolor{gray!25}0.25 &
\cellcolor{blue!25}0.39 & \cellcolor{red!25}0.50 & \cellcolor{gray!25}0.24 &
\cellcolor{blue!25}0.30 & \cellcolor{red!25}0.31 & \cellcolor{gray!25}0.17 &
\cellcolor{blue!25}0.18 & \cellcolor{red!25}0.46 & \cellcolor{gray!25}0.28 
\\
    \hline
p@10 &
\cellcolor{blue!25}0.39 & \cellcolor{red!25}0.48 & \cellcolor{gray!25}0.30 &
\cellcolor{blue!25}0.44 & \cellcolor{red!25}0.56 & \cellcolor{gray!25}0.26 &
\cellcolor{blue!25}0.33 & \cellcolor{red!25}0.38 & \cellcolor{gray!25}0.20 &
\cellcolor{blue!25}0.19 & \cellcolor{red!25}0.50 & \cellcolor{gray!25}0.30\\ 
\hline    
    \end{tabular}

\end{center}
\caption{{Evaluating the importance of topical cohesion. Blue, red and gray denote Europarl, Wikipedia and a mixed corpus where English is sampled from Europarl and the other language (Spanish or German) is sampled from Wikipedia, respectively. Results indicate that lack of topical cohesion affects performance. However, in spite of reduced topical cohesion, our method still retrieves bilingual lexicons of reasonable size.}}
\label{tab:cohesion}}
\end{table*}

\subsection{Translation retrieval task}

We evaluated our approach on a task of translation retrieval. We followed the same experimental protocol described in~\cite{conneau2017word}. We used 300K English sentences and 300K Italian sentences from the Europarl corpus to learn the bilingual lexicons using our constrained nearest neighbor sampling method. The translation retrieval task involves mapping 2k sentences from a source language to 2k in the target language out of a pool of 200K sentences in the target language. For a sentence in the source language, \texttt{Method 2} uses the \texttt{translateEmbeddings} algorithm as described in Algorithm~\ref{algo:NN-sampling} and obtains an equivalent (noisy) embedding in the target language. Next, using the obtained embedding, it finds the nearest neighbor(s) in the 200K sentences in target language ranked by cosine distance in the ascending order. In \texttt{Method 1}, we perform a minor modification in \texttt{translateEmbeddings}. Recall that, in \texttt{translateEmbeddings}, for a given source word, the Algorithm first finds potential translations, and then prunes it to only include translations that include the source word when we translate back; from this pruned set, we randomly pick a target word. Unlike the previous task of detecting \emph{hope speech} where we intended to retrieve a diverse pool of comments, in this task, we are more interested in finding the exact match of a source sentence in a target language. For this reason, instead of randomly sampling from the pruned list of target word choices for a source word, we always select the top choice. 

Our performance is summarized in Table~\ref{tab:translationRetreival}. Understandably, our methods are outperformed by existing sophisticated methods that perform explicit alignment. For \emph{en} $\rightarrow$ \emph{it} translation, we obtained slightly better performance than~\cite{mikolov2013exploiting}. In contrast with \cite{mikolov2013exploiting}, our method conducts no explicit attempt to align and does not require any seed lexicon (\cite{mikolov2013exploiting} requires a lexicon of 5,000 words).  

\renewcommand{\tabcolsep}{1mm}
\begin{table}[t]
\scriptsize
{

\begin{center}
     \begin{tabular}{|l | c  | c | c| c| c | c|}
     \hline 
      & \multicolumn{3}{c|}{\emph{en} $\rightarrow$\emph{it}}   & \multicolumn{3}{c|}{\emph{it} $\rightarrow$\emph{en}}  \\
    \hline
Methods &    p@1 &  p@5 &  p@10 & p@1 & p@5 & p@10\\
\hline
\cite{mikolov2013exploiting} &0.11 &0.19 &0.23 &0.12 &0.22 &0.27\\
\hline 
\cite{dinu2014improving} &0.45 &0.72 &0.81 &0.49 &0.71 &0.78\\
\hline 
\cite{smith2017offline} &0.55 &0.73 &0.78 &0.43 &0.62 &0.69\\
\hline 
\cite{conneau2017word} &0.66 &0.80 &0.83 &0.69 &0.80 &0.83\\
\hline
\texttt{Method 1} &0.16 &0.27 &0.32 &0.03 &0.06 &0.08\\
\hline
\texttt{Method 2} &0.12 &0.22 &0.36 &0.03 &0.06 &0.08\\
\hline
    \end{tabular}

\end{center}
\caption{{Evaluating translation retrieval performance. We follow the evaluation task as presented in~\cite{conneau2017word}. The translation task involves mapping 2K randomly sampled sentences in a source language to 200K sentences in the target language.}}
\label{tab:translationRetreival}}
\end{table}

\newpage

\end{document}